\documentclass{midl} 
\usepackage{graphicx} 
\usepackage{framed}
\definecolor{shadecolor}{RGB}{0,0,0}
\usepackage{comment}
\newlength\savewidth\newcommand\shline{\noalign{\global\savewidth\arrayrulewidth
		\global\arrayrulewidth 1pt}\hline\noalign{\global\arrayrulewidth\savewidth}}

\usepackage{amsmath,amssymb,amsfonts,caption}
\usepackage{algorithmic}
\usepackage{graphicx}
\usepackage{textcomp}
\usepackage{multirow,booktabs}

\usepackage{multirow}
\usepackage{graphicx}
\usepackage{wrapfig}
\newcommand{\bumpup}{\vspace*{-2.0ex}}

\jmlryear{2025}\jmlrworkshop{Full Paper -- MIDL 2025}\jmlrvolume{-- 141}\editors{Accepted for publication at MIDL 2025}

\title[DAGMaN: Distilled smart-attention with noisy teacher]{Co-distilled attention guided masked image modeling with noisy teacher for self-supervised learning on medical images}

\midlauthor{
	\Name{Jue Jiang\nametag{$^{1}$}} \Email{jiangj1@mskcc.org}\\
	\Name{Aneesh Rangnekar\nametag{$^{1}$}} \Email{rangnea@mskcc.org}\\
	\Name{Harini Veeraraghavan\nametag{$^{1}$}} \Email{veerarah@mskcc.org}\\
	\addr $^{1}$ Memorial Sloan Kettering Cancer Center
}

\begin{document}
	
	\maketitle
	
	\begin{abstract}
		Masked image modeling (MIM) is a highly effective self-supervised learning (SSL) approach to extract useful feature representations from unannotated data. Predominantly used random masking methods make SSL less effective for medical images due to the contextual similarity of neighboring patches, leading to information leakage and SSL simplification. \textcolor{black}{Hierarchical shifted window (Swin) transformer, a highly effective approach for medical images cannot use advanced masking methods as it lacks a global [CLS] token. Hence, we introduced} an attention guided masking mechanism for Swin within a co-distillation learning framework to selectively mask semantically co-occurring and discriminative patches, to reduce information leakage and increase the difficulty of SSL pretraining. However, attention guided masking inevitably reduces the diversity of attention heads, which negatively impacts downstream task performance. To address this, we \textcolor{black}{for the first time}, integrate a noisy teacher into the co-distillation framework (termed DAGMaN) that performs attentive masking while preserving high attention head diversity. We demonstrate the capability of DAGMaN on multiple tasks including full- and few-shot lung nodule classification, immunotherapy outcome prediction,  tumor segmentation, and unsupervised organs clustering.
	\end{abstract}
	
	
	\section{Introduction}
	Self-supervised learning (SSL) is an approach to extract useful feature representations from large cohorts of unlabeled images in a task agnostic way, following which the model can be applied to downstream tasks with minimal fine-tuning. Masked image modeling (MIM) is a highly effective SSL task that extracts useful feature representations from degraded images~\cite{he2021masked,zhou2022image,jiang2022self_SMIT}. MIM fosters token diversity across attention heads and increases locality inductive bias for transformers, which enhances feature reusability for downstream tasks~\cite{XieCVPR2023_SecretsMIM}. MIM is often implemented using random~\cite{zhou2022image,jiang2022self_SMIT} or blockwise~\cite{bao2021beit} masking of image patches. However, such methods are unsuitable for medical images due to the strong semantic correlation of spatially adjacent patches, which risks information leakage and simplification of the SSL task. Recent work has focused on making MIM challenging by selectively masking discriminative regions. Strategies include masking high-attending~\cite{kakogeorgiou2022hide_attmask,LiuAAAI2023_AttentionThrow}, low-attending~\cite{li2021mst}, as well as semantically relevant~\cite{li2022semmae,shi2022adversarial_mask} and hard to reconstruct regions~\cite{Wang_2023_CVPR} derived from auxiliary networks for natural images. The impact of attention guided MIM has not been studied for medical image analysis. 
	Hierarchical shifted window (Swin) transformers enable multi-scale attention with linear complexity, making them ideal for analyzing volumetric medical images~\cite{tang2022self}. Hybrid Swin-convolutional segmentation models combining Swin encoder with convolutional decoders have demonstrated higher accuracy compared to ViT and convolutional networks~\cite{tang2022self,jiang2024,cao2022swin}. However, attentive masking cannot be performed with Swin due to two key architectural limitations. First, Swin uses windowed attention that limits attention to local regions. Second, Swin lacks the classifier token [CLS] that interacts with all the patch embeddings via self-attention to extract global attention required for attention guided masking. 
	Hence, \textcolor{black}{we introduced an architectural enhancement to Swin} for performing attention guided MIM. Our \textcolor{black}{enhancement is a semantic attention (SA) module}, consisting of vision transformer (ViT) blocks implemented into Stage\#3 with a [CLS] token \textcolor{black}{interact with token embeddings across the whole image}. Similar to AttMask~\cite{kakogeorgiou2022hide_attmask}, an exponentially moving average (EMA) teacher guides the masking applied to a co-distilled student network. Unlike AttMask, our approach works for Swin and ViT. Furthermore, to enhance attention diversity~\cite{arani_FickleTeacherWACV2021}, \textcolor{black}{we introduced a noisy teacher} into the co-distillation learning framework by applying patch dropout to the teacher's input. We call our approach that performs co-Distillation Attention Guided MAsking with Noisy teacher, DAGMaN. We show DAGMaN produces global self-attention and achieves higher attention head diversity.
	Our contributions are: (i) architectural enhancement of Swin to perform attentive masking, (ii) noisy teacher regularized co-distillation for enhancing attention-head diversity, (iii) analysis of the impact of our DAGMaN approach on multiple medical image analysis tasks including lung nodule classification, segmentation of nodules and malignant tumors, lung cancer treatment response prediction, and unsupervised clustering to distinguish organs. All the code and model checkpoints will be made available upon manuscript acceptance. 
	
	\bumpup
	\bumpup
	\section{Related works}
	\textbf{Masking strategies for MIM \/:}
	Random masking~\cite{he2021masked,jiang2022self_SMIT,xie2021simmim,zhou2022image} is a commonly used MIM approach due to its computational simplicity. A key limitation of this approach is the potential of exposing highly correlated and spatially adjacent patches, which simplifies SSL and reduces downstream task accuracy. Block-wise matching~\cite{wei2022masked_feature, bao2021beit, wang2022bevt} mitigates this issue by masking spatially adjacent patches. Masked regions are still selected randomly, dispersing attention towards irrelevant regions, and wasting computations for pretraining.
	\\
	Attentive masking methods overcome the aforementioned issues by identifying and masking discriminative and spatially coherent regions. Semantic attention~\cite{li2022semmae} and hard patch selection methods~\cite{Wang_2023_CVPR} use auxiliary networks to identify discriminative regions that add computational complexity and need for different pretrained networks. ViT-based methods use the spatial attention maps computed using the [CLS] token to identify regions for masking, providing a computationally simple approach~\cite{li2021mst,kakogeorgiou2022hide_attmask,LiuAAAI2023_AttentionThrow}. However, attention guided masking with [CLS] is not possible to perform with Swin, which our approach overcomes.
	\\
	\textbf{Noisy \textcolor{black}{teacher} co-distillation \/:}
	Noise injection is an implicit data augmentation method~\cite{you2022simcvd} that improves data utilization with reduced memory requirements and improves training convergence~\cite{liu2023patchdropout}. \textcolor{black}{Feature} dropout-based noise injection has been used for knowledge distillation applied to \textcolor{black}{few or feature layers using same}~\cite{lee2023self} as well as \textcolor{black}{different teacher and student models for classification tasks}~\cite{arani_FickleTeacherWACV2021,liu2020noisy,bulo2016dropout}. \textcolor{black}{Distillation with noise injected into intermediate outputs or features} has shown to enhance accuracy for image-level~\cite{tarvainen2017mean, laine2016temporal} and dense-pixel prediction tasks~\cite{you2022simcvd}. \textcolor{black}{Different from prior methods, we inject noise into teacher's inputs to increase the variability of extracted features by the two networks, that in turn enhances attention head diversity.} 
	
	\begin{figure*}[t]
		\centering
		\includegraphics[trim=20 10 3 12, clip,width=1.0\textwidth]{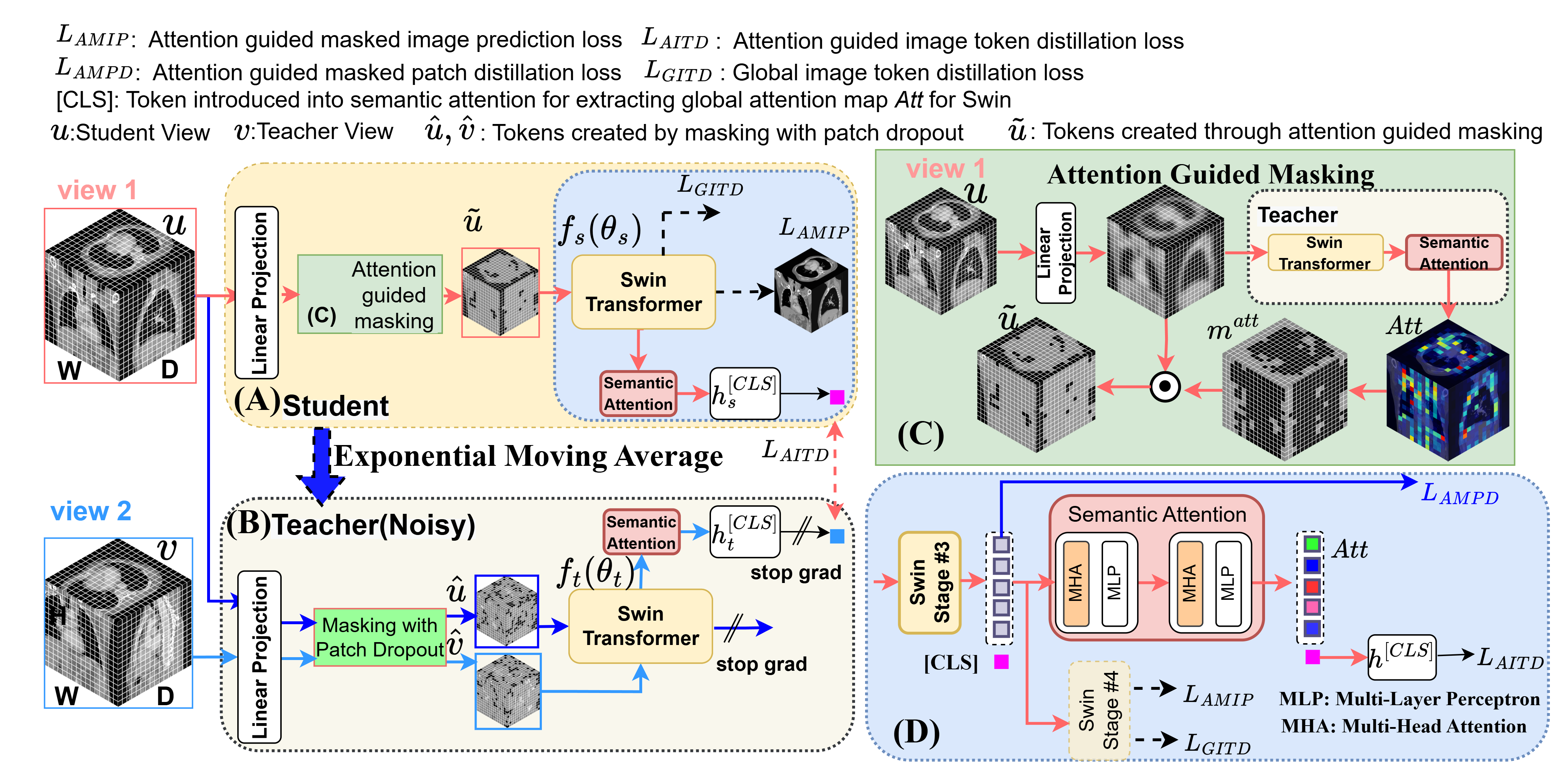}
		\caption{DAGMaN: (A) shows the student network with input produced using attention guided masking. (B) shows the noisy teacher that is provided noisy tokens produced using patch dropout. \textcolor{black}{(C) shows the attention guided masking produced using global attention mask $Att$ produced by the teacher. (D) depicts the semantic attention block with [CLS] token implemented into Swin in order to extract $Att$.}}
		\label{fig:method}
	\end{figure*}
	\bumpup
	\section{Method}
	Swin cannot extract global attention because it conducts self-attention locally within defined spatial windows and shifted window partitioning to maintain cross-window relationships. \textcolor{black}{We introduced semantic attention (SA) block into Swin to extract global attention for attention-guided masking and a noisy teacher co-distillation for attention diversification.}  
	
	\bumpup
	
	\subsection{Semantic attention guided masked image modeling}
	\textcolor{black}{SA (Figure~\ref{fig:method} D) is composed of 2 ViT blocks, with each block consisting of a multi-head attention layer followed by a Multi-Layer Perceptron Layer (MLP)}. A [CLS] token $z_{[CLS]}=z_{N+1}$ is introduced at SA, and interacts with all $i \in N$ features $\{z_{i}\}_{i=1}^{N}$ and creates a tokenized feature z $\in$ $\mathbb{R} ^{(N+1)\times D}$, where $D$ is the feature size. The multi-head mechanism using $h$ attention heads and individual projection matrices $W_{q}$, $W_{k}$, $W_{v}$ extracts various representations using query $Q^{[CLS]} = W_{q} \times z_{[CLS]} + b_{q}$, key $K = W_{k} \times z + b_{k}$, and value $V = W_{v} \times z + b_{v}$ features, where $b_q, b_k, b_v$ are the biases. The semantic attention (S$_{ATT}$) averages the representations as:
	\begin{equation}
		\setlength{\abovedisplayskip}{1pt}
		\setlength{\belowdisplayskip}{1pt} 
		S_{ATT} = \frac{1}{h} \sum_{1}^{h} \mathrm {Softmax}(Q_h^{[CLS]} \cdot \frac{K_h^T}{\sqrt{D/h}}). 
		\label{eqn:attmsk}
	\end{equation}  
	$S_{ATT}$ $\in$ [0,1] is an attention vector of the same size $N$ as the number of patches, with larger values indicating higher attention. Thus, S$_{ATT}$ is sorted in descending order to identify the top [$rN$] patches, $r \in [0,1]$ being the masking ratio, default to 0.7, and setting those indices in mask vector $m^{satt}$ to 0. Inputs to $f_s(\theta_{s})$ is produced by $\tilde{u} = m^{satt} \odot u$ (Figure~\ref{fig:method} C). \\
	Masking all the high-attending tokens can remove the most discriminatory regions, potentially making the SSL task very challenging and hamper convergence. Hence, a small number $s < r$ (default of 0.1), of the top high attending or hint tokens are left unmasked. \\
	\textcolor{black}{Swin adopts a hierarchical design, as a result of which features extracted at earlier stages (\#1 and \#2) may not contain sufficient global context. On the other hand, features extracted after stage \#4 have very low resolution. Hence, stage \#3 was selected for SA module to balance network depth and attention density.} 
	
	\bumpup
	\subsection{\textcolor{black}{Attention guided }Co-distillation framework \textcolor{black}{with Noisy teacher}}
	\label{subsec:noisyteacher}
	Two different cropped views ($u$ and $v$) from the input 3D CT volume are converted into a sequence of $N$ patch tokens $\{u_{i}\}_{i=1}^{N}, \{v_{i}\}_{i=1}^{N}$ and then subjected to guided and random masking to create corrupted input patch tokens $\tilde{u}$ and $\hat{v}$ for student ($f_{s}(\theta_{s})$ and teacher ($f_{t}(\theta_{t})$) networks, respectively (Figure~\ref{fig:method}). The networks' parameters are updated using co-distillation losses consisting of  attention guided image token distillation (AITD), attention guided masked patch token distillation (AMPD), global image token distillation (GITD). The attention guided masked image prediction (AMIP) is used only for the student model. \textcolor{black}{\(L_{AITD}\) measures dissimilarity of the token embeddings with [CLS] and optimizes SA.\(L_{AMPD}\) measures the dissimilarity of patch token embeddings extracted by teacher and student networks, emphasizing dissimilarities in the attention guided masked regions. AITD and AMPD losses are computed after stage \#3 to take advantage of the denser feature representation compared to stage \#4 and higher anatomic context than earlier stages. \(L_{GITD}\) that measures the dissimilarity of the token embeddings computed for the whole image views and \(L_{AMIP}\) that measures difference in the generated image patches in the masked regions are computed at the end of stage \#4.}
	
	\noindent\textbf{Noisy teacher regularization\/ }\rm  The teacher network (Figure.~\ref{fig:method}B) uses an identical architecture as the student and is created using exponential moving average as $\theta_{t}=\lambda_m\theta_{t} + (1-\lambda_m)\theta_{s}$, where $\lambda_{m}$ is momentum. \textcolor{black}{Diversity of attention head is increased by using} patch dropout~\cite{liu2023patchdropout} to create noisy input tokens $\hat{u}, \hat{v}$ from $u,v$ for the teacher, thus termed noisy teacher. Patch dropout randomly sets elements within mask tokens $m_{u},m_v$ corresponding to indices of patch tokens $u,v$ to 0 using a patch drop ratio $r_t \in [0,1]$, where $r_t = 0.7$. The corrupted inputs $\hat{u} = m_u \odot u$ and $\hat{v} = m_v \odot v$ are thus generated for $f_{t}(\theta_{t})$.  \\
	\noindent\textbf{Attention guided image token distillation \/}\rm loss measures the dissimilarity of the token distributions $P_{s}^{[CLS]} (\tilde{u},\theta_{s})$ and $P_{t}^{[CLS]}(\hat{v},\theta_{t})$, \textcolor{black}{computed for $\tilde{u}$, $\hat{v}$}, produced by linear projection layers placed after the SA blocks in the student $f_{s}(\theta_{s})$ and teacher $f_{t}(\theta_{t})$ networks:
	\begin{equation}
		\setlength{\abovedisplayskip}{1pt}
		\setlength{\belowdisplayskip}{1pt} 
		\begin{split}
			L_{AITD} = - \sum_{i=1}^N P_{t}^{[CLS]}(\hat{v}_{i},\theta_t) log (P_{s}^{[CLS]}(\tilde{u}_{i},\theta_s)).
		\end{split}
	\end{equation}
	The sharpness of the token distribution is controlled by computing a sharpening transformation~\cite{jiang2022self_SMIT} using separate temperature terms $\tau_{s} >0$ and $\tau_{t} > 0$ for the student and teacher networks. 
	\noindent\textbf{Attention guided masked patch distillation \/}\rm loss measures the dissimilarity of masked token distributions $P_t^{Patch}(\hat{u}, \theta_t)$ and $P_s^{Patch}(\tilde{u}, \theta_{s})$, produced from the different corrupted versions (\textcolor{black}{$\hat{u}$ from patch dropout, $\tilde{u}$ from attention guided masking}) of same view $u$ to increase robustness of network to image noise. The token distributions are computed by applying \textit{softmax\/} operation to the stage \#3 output of $f_{t}(\theta_{t})$ and $f_{s}(\theta_{s})$, followed by sharpening transformations \textcolor{black}{and averaged on the $m_i^{att}$ tokens.} AMPD is computed as:  
	\begin{equation}
		\setlength{\abovedisplayskip}{1pt}
		\setlength{\belowdisplayskip}{1pt} 
		\begin{split}
			L_{AMPD} = - \sum_{i=1}^N m_i^{att} \cdot  P_{t}^{Patch}(\hat{u}_{i},\theta_t) log (P_{s}^{Patch}(\tilde{u}_{i},\theta_s)).
		\end{split}
	\end{equation}
	\\
	
	\noindent\textbf{Global image token distillation \/}\rm loss measures the dissimilarity of the global image token distributions $P_{s}^{[g]}(\tilde{u},\theta_{s})$ and $P_{t}^{[g]}(\hat{v},\theta_{t})$ produced from the average pooling layer placed after Stage \#4, \textcolor{black}{computing the mean of all embeddings in the spatial dimension, producing a global feature representation}  of the student and teacher networks, respectively. $I_{GITD}$ is then computed as:
	\begin{equation}
		\setlength{\abovedisplayskip}{1pt}
		\setlength{\belowdisplayskip}{1pt} 
		\begin{split}
			L_{GITD} = - \sum_{i=1}^N P_{t}^{[g]}(\hat{v}_{i},\theta_t) log (P_{s}^{[g]}(\tilde{u}_{i},\theta_s)).
		\end{split}
	\end{equation}
	The token distributions are subjected to sharpening transforms as used for AITD. 
	\\
	
	\noindent\textbf{Attention guided masked image prediction \/}\rm measures the image reconstruction error between the reconstructed image  produced by $f_{s}(\theta_{s})$ using a 1-layer linear projection layer $h_{s}^{Pred}$ applied to masked input $\tilde{u}$ as: 
	\begin{equation}
		\setlength{\abovedisplayskip}{1pt}
		\setlength{\belowdisplayskip}{1pt} 
		L_{AMIP} = \sum_{i}^{N} E\| m_{i}^{att} \cdot (h_{s}^{Pred}(f_{s}(\tilde{u_{i}}, \theta_{s})) - u_{i}) \|_{1}, 
	\end{equation} 
	where $m_{i}^{att}$ is the masked token vector produced through attention guided masking.  The total loss is computed as, $L_{total}$ = $L_{AMIP}$ + $\lambda_{AMPD}$ $L_{AMPD}$ + $\lambda_{AITD}$ $L_{AITD} + \lambda_{GITD} G_{ITD}$.

	\bumpup
	\section{Experiments and Results}
	
	\subsection{Datasets}
	\label{subsec:pretrainingdatasets}
	\textbf{Pretraining: \/}\rm SSL pretraining was performed on  10,412 unlabeled 3D CTs (1.89M images) sourced from public~\cite{xiao2023lesion,C4KC-KiTS,ji2022amos} and institutional cohorts involving diseases in the head and neck, chest, and abdomen~\cite{jiang2024}.
	\\
	\textbf{Downstream tasks and data: \/}\rm A total of 4,746 cases sourced from three public~\cite{armato2011lung,aerts2015data,yang2023medmnist} and one institutional dataset~\cite{jiang2024} were evaluated for lung nodule classification (Task 1) using lung image database consortium (LIDC)~\cite{armato2011lung}, lung nodule and tumor segmentation (Task 2), immunotherapy response prediction (Task 3)~\cite{zhu2023wasserstein}, and unsupervised clustering of multiple organs using pretrained features (Task 4). Description of the various datasets and fine-tuning details are included in Supplementary Section~\ref{sup_subsec:downsteamdatadescriptions}. 
	
	\subsection{Experiments}
	\label{subsec:finetuningdatasets}
	Implementation details for the various networks and analyzed tasks are provided in the Supplemental Section~\ref{sup_subsec:implementationdetails}. Experiments evaluated the benefit of attention guided masking and noisy teacher regularization with Swin. Hence, DAGMaN was compared against SMIT~\cite{jiang2022self_SMIT}, which uses random masking with an identical Swin backbone as well as blockwise masking as used in iBot~\cite{zhou2022image}. ViT-based AttMask~\cite{kakogeorgiou2022hide_attmask} as well as low attending masking used in masked self-supervised transformer (MST)~\cite{li2021mst} were also analyzed. All networks were pretrained using identical pretraining, fine-tuning, and testing datasets. Few-shot performance using 25\% and 50\% cases was used to assess classification (Task 1, Task 3) and segmentation (Task 2). \textcolor{black}{Representative CNN baselines for segmentation, nnU-Net\cite{isensee2021nnu} and classification, 3D-ResNet 50\cite{chen2019med3d} were used for lung nodule segmentation as well as lung nodule malignancy and immunotherapy outcome prediction, respectively.}  
	
	\subsection{Image-level classification performance (Task 1 and 3)}
\begin{figure}[t]
		\centering
		\includegraphics[width=0.95\linewidth]{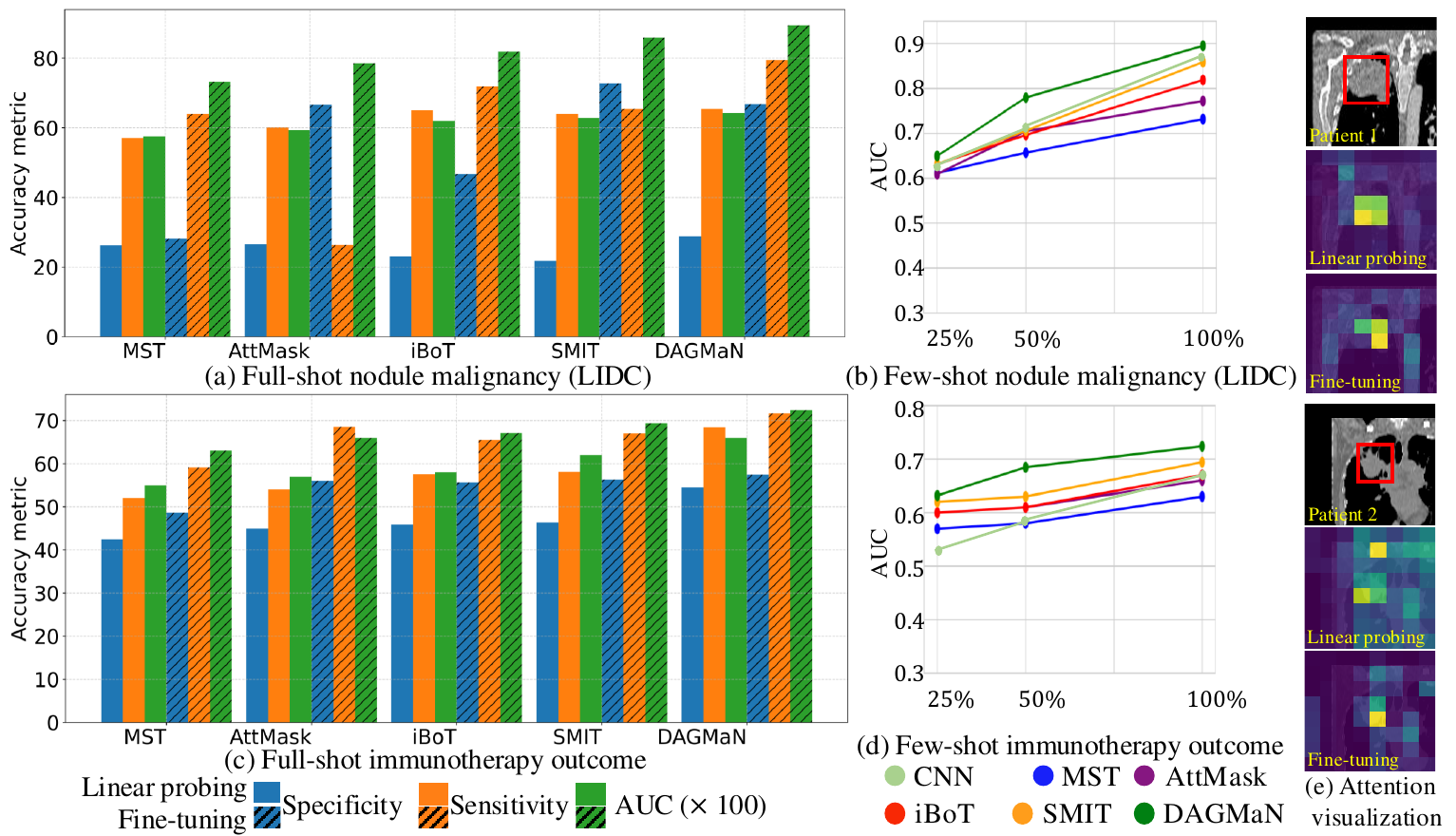}
		\setlength{\belowcaptionskip}{-0.3cm}
		\caption{Accuracy comparison of various methods \textcolor{black}{including CNN-based 3D ResNet-50 } on two different classification tasks using full-shot (a,c) and few-shot (b,d) training regimes. (e) shows attention maps computed with LP and FT for two representative lung cancer patients, who did not respond to treatment.}
		\label{fig:classification}
	\end{figure}

	Image-level classification performance was evaluated by using linear probing (LP) with one projection head and fine-tuning (FT) by training all the layers for both Task 1 and 3. DAGMaN was more accurate than other methods, resulting in higher AUC using both FT and LP (Figure~\ref{fig:classification}). Few-shot training showed higher accuracy with DAGMaN than other methods even with 25\% training data. \textcolor{black}{3D ResNet was the second best method for predicting lung nodule malignancy even at but produced worst accuracy with 25\% training data for predicting immunotherapy outcomes.} DaGMaN had higher specificity and sensitivity compared to other methods for nodule malignancy classification (Task 1) but similar specificity as other methods for predicting immunotherapy outcome (Task 3). The attention maps computed for two different patients with LP and FT using DAGMaN clearly show localization within the tumor (Figure~\ref{fig:classification} (e)).   
	
	\subsection{Dense-pixel prediction as tumor segmentation (Task 2)}
\begin{figure}
		\centering
		\includegraphics[width=1.00\linewidth]{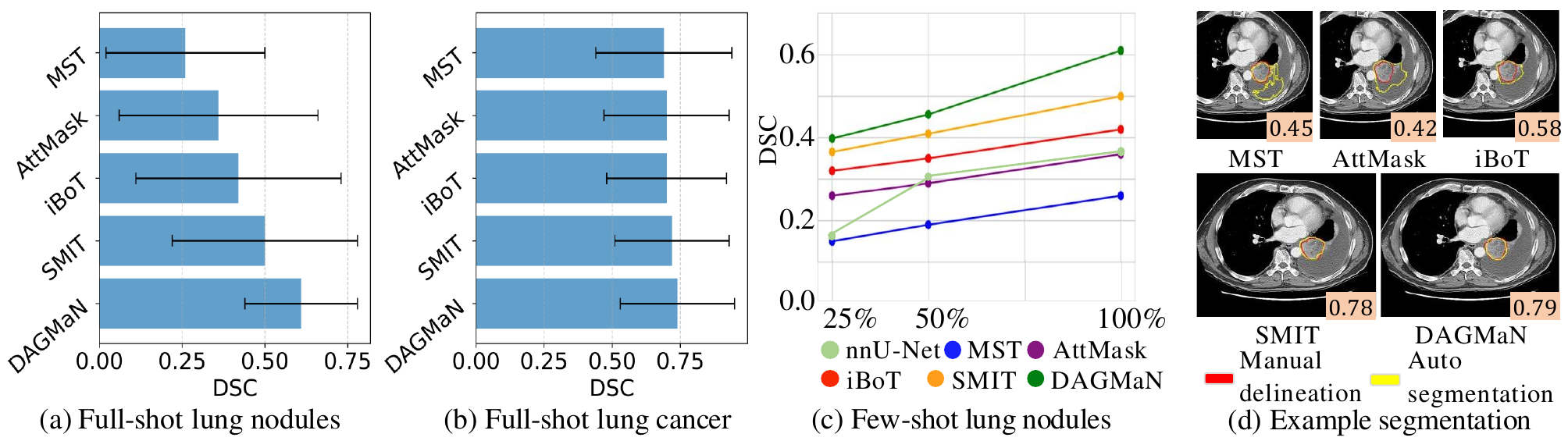}
		\setlength{\belowcaptionskip}{-0.4cm}
		\caption{Full-shot segmentation for (a) lung nodules, (b) tumors, as well as (c) few-shot performance for lung nodules. (d) shows a representative patient with DSC accuracies.}
		\label{fig:segmentation}
	\end{figure}

	The segmentation network was implemented by adding a U-Net decoder, initialized from scratch, to the pretrained encoders and all networks subjected to FT~\cite{aerts2015data} (training with 350 and validation with 27 cases). Results of testing the model with LIDC and immunotherapy datasets are shown for full-shot (Figure~\ref{fig:segmentation} (a,b)) and few-shot (Figure~\ref{fig:segmentation} (c)) regimes. DAGMaN outperformed all methods in both full- and few-shot regimes and showed clear performance improvement for the nodule segmentations compared to segmenting larger lung tumors. \textcolor{black}{nnUnet was less accurate than all except MST under all analyzed training data settings, indicating the importance of pretext task in SSL pretraining. It was also similar in performance to AttMask, which uses a ViT architecture.} Segmentations for a representative case is shown in Figure~\ref{fig:segmentation} (d). 
	\subsection{Impact of attentive masking and noisy teacher co-distillation}
\begin{figure}[ht]
		\centering
		\includegraphics[trim=55 75 75 28, clip,width=1.00\linewidth]{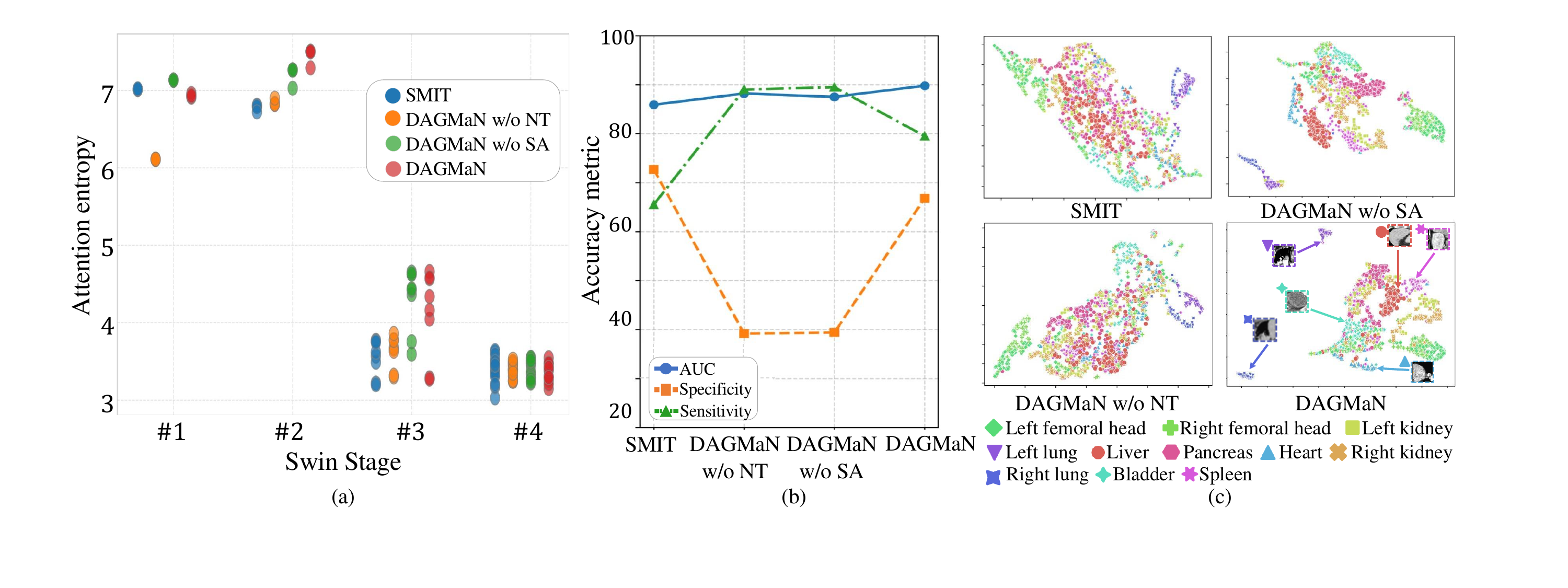}
		\setlength{\belowcaptionskip}{-0.4cm}
		\caption{Impact of attentive masking and noisy teacher co-distillation. (a) shows attention-head entropy across the Swin stages with different masking strategies \textcolor{black}{on the LIDC dataset}. (b) shows the impact of the same strategies on lung nodule malignancy classification and (c) shows differences in unsupervised clustering of distinct organs using pretrained features.}
		\label{fig:attention_diversity_quantitative}
	\end{figure}

	\textcolor{black}{Diversity of attention heads measured using entropy of attention distances} for all the 4 Swin stages is shown in Figure~\ref{fig:attention_diversity_quantitative} (a). Random masking as performed in SMIT resulted in a high entropy with a small variance in the first two stages but lower entropy with increasing variance in the last two stages. \textcolor{black}{Using only semantic attention (SA)} reduced average entropy and variance in stages 1 and 4, \textcolor{black}{indicating reduced diversity. On the other hand}, adding noisy teacher (NT) increased entropy and variance even for random masking (stages 2 and 3). Similarly, DAGMaN that uses both SA and NT produced high entropy in stages 1, 2, and 3 with higher variance than other methods indicating higher attention diversification. 
	\\
	The impact of attention diversification was analyzed for lung nodule malignancy classification (Figure~\ref{fig:attention_diversity_quantitative} (b)), which showed an overall higher AUC with DAGMaN compared to all other methods. Training using only SA or NT produced substantially lower specificity than DAGMaN, indicating that combining NT and SA improves accuracy. \\
	UMAP clustering~\cite{mcinnes2018umap} performed using features computed from the pretrained models on the OrganMNIST3D dataset showed the best separation of various organs (Task 4) with DAGMaN (Figure~\ref{fig:attention_diversity_quantitative} (c)) with highest inter-class and lowest intra-class distance (Supplemental Figure 7). SA guided masking improved organ separation compared to random masking (SMIT) and NT, but was less effective than DAGMaN. 
	
	\subsection{Ablation tests}
	Analysis of various pretraining losses for nodule malignancy classification (Task 1) showed that removing attentive masked image prediction (AIMP) resulted in the largest accuracy drop (AUC of 0.829) followed by attentive masked patch distillation (AMPD) with AUC of 0.865 (Supplemental Table \ref{tab:ablation_differentlosses}). GITD least impacted accuracy indicated by a small drop in AUC of 0.878. However, inclusion of GITD with other losses produced best AUC of 0.895. \\
	Figure.~\ref{fig:ablation_attnmap_differentstages_samodule_placement} depicts the impact of placing SA module in the different stages of the Swin encoder. As shown, SA placed in stage \#3 resulted in the highest accuracy with both linear probing and fine-tuning for classifying malignancy of lung nodules. Similarly, SA placed in stage \#3 also resulted in highest accuracy for segmenting the lung nodules (DSC of 0.9 with FT, DSC of 0.7 with LP) compared to all other stages (second highest AUC of 0.7 for stage \#4 with FT). SA placed in stage \#3 also resulted in the highest inter-cluster and lowest inter-cluster separation for Task 4 (Supplemental Figure \ref{fig:ablation_clustering_samodule_placement}). 
	\\
	Finally, we implemented DAGMaN in ViT and found that it achieved higher accuracy for Task 1 (AUC of 0.66 with LP and 0.847 with FT) compared to random masking (AUC of 0.643 with LP and 0.816 with FT) as shown in Supplemental Table \ref{tab:ablation_performance_vitswin}.
	
	\bumpup
	\subsection{DAGMaN learns better attention maps}
\begin{figure}[t]
		\centering
		\includegraphics[width=0.8\linewidth]{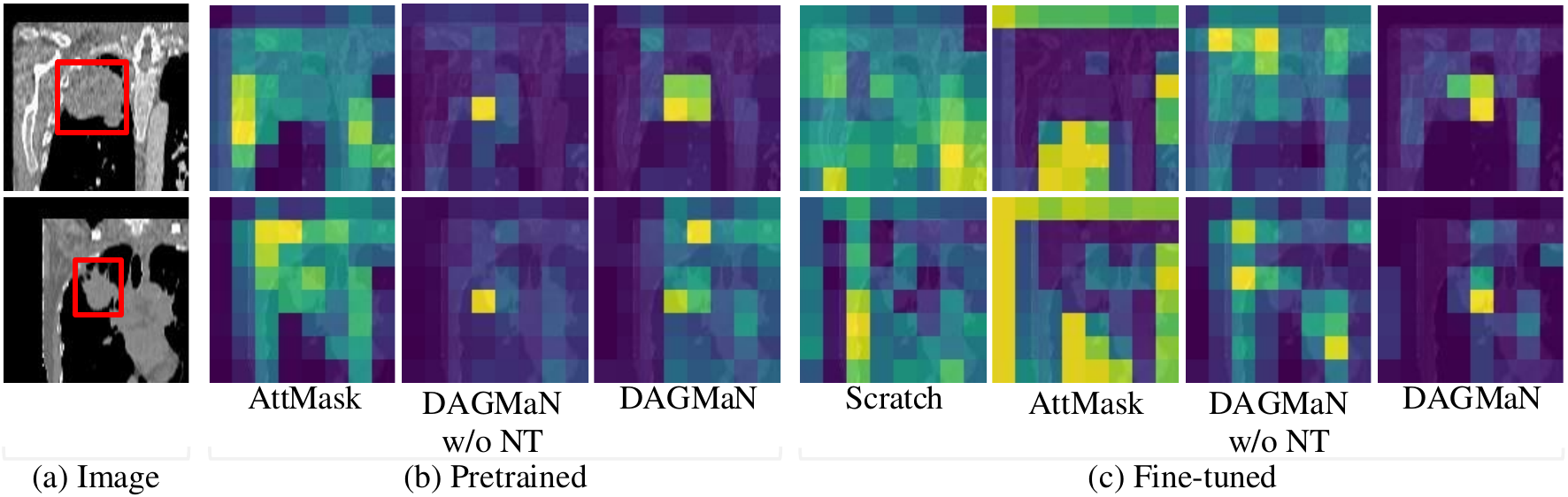}
		\caption{Attention maps for representative examples from the immunotherapy response prediction dataset. (a) shows CT image slice. Attention maps are shown for (b) semantic attention layer pretrained with AttMask, DAGMaN w/o noisy teacher and DAGMaN, (c) semantic attention layer fine tuned with pretrained weights, including training from scratch, AttMask, DAGMaN w/o noisy teacher and DAGMaN.}
		\label{fig:attention_diversity_qualitative}
	\end{figure}

	We analyzed the capability of DAGMaN to extract attention maps for classification task 3 involving predicting immunotherapy outcome. DAGMaN was evaluated against AttMask~\cite{kakogeorgiou2022hide_attmask} as well as without noisy teacher. Our analysis showed that DAGMaN produced attention visualizations focused on the tumor both using pretrained features (Figure~\ref{fig:attention_diversity_qualitative} (b)) and following fine-tuning (Figure~\ref{fig:attention_diversity_qualitative} (c)). AttMask resulted in activations everywhere including lungs and chest, whereas noisy teacher co-distillation without semantic attention resulted in a dispersed attention, indicating importance of semantic attention.

	\section{Discussion and conclusion}
	Swin is a more accurate architecture than ViT for medical applications~\cite{tang2022self,jiang2024,cao2022swin}. 
	We introduced an approach to extract attention maps from Swin, which for the first time allows attention guided masking as well as capability to visualize explanation of model's prediction for classification tasks. We also introduced a noisy teacher co-distillation approach, which shows that combining semantic attention guided masking with noisy teacher enhances the attention head diversity in the first three stages of Swin while increasing downstream accuracy in both full- and few-shot training regimes for image-level and dense-pixel prediction tasks. Our results are consistent with observations for natural images that also showed increased attention diversity with masked image modeling improves accuracy~\cite{XieCVPR2023_SecretsMIM}. DAGMaN produced robustly accurate segmentations independent of tumor size variations. Although we developed our approach for Swin, we found that DAGMaN was also applicable to ViT. \textcolor{black}{DAGMaN though more resilient than other methods, was adversely impacted by tumors occurring in the mediastinum, as well as tumors fused with collapsed lung. Both of these are difficult conditions for segmentation.} Future work will also focus on combining DAGMaN-based approach for parameter efficient fine-tuning by understanding the dynamics of feature refuse, which is beyond the scope of this work. 
	
	\section{Acknowledgements} 
	This research is the result of funding partially supported by the NCI R01CA258821 and the Memorial Sloan Kettering Cancer Center Support Grant/Core Grant NCI P30CA008748. 
	
	\clearpage
	\bibliography{midl25_141}

@misc{C4KC-KiTS,
 title={C4KC-KiTS},
 author={Heller, N. and Sathianathen, N. and Kalapara, A. and Walczak, E. and Moore, K and Kaluzniak, H and Rosenberg, J and Blake, P and Rengel, Z and Oestreich, M and Dean, J and Tradewell, M and Shah, A and Tejpaul, R and Edgerton, Z and Peterson, M and Raza, S and Regmi, S and Papanikolopoulos, N and Weight, C.},
 howpublished={The Cancer Imaging Archive.},
 year=2019
}

@INPROCEEDINGS{Wang_2023_CVPR,
  author={Wang, Haochen and Song, Kaiyou and Fan, Junsong and Wang, Yuxi and Xie, Jin and Zhang, Zhaoxiang},
  booktitle={2023 IEEE/CVF Conference on Computer Vision and Pattern Recognition (CVPR)}, 
  title={Hard Patches Mining for Masked Image Modeling}, 
  year={2023},
  pages={10375-10385}
}

@article{jiang2024,
    author = {Jiang, J and Rangnekar, A and Veeraraghavan, H},
    title = {Self-supervised learning improves robustness of deep learning lung tumor segmentation models to CT imaging differences},
    journal = {Med Phys},
    year = {2024}
}

@inproceedings{he2021masked,
  title={Masked autoencoders are scalable vision learners. 2022 IEEE},
  author={He, Kaiming and Chen, Xinlei and Xie, Saining and Li, Yanghao and Doll’ar, Piotr and Girshick, Ross B},
  booktitle={CVF Conference on Computer Vision and Pattern Recognition (CVPR)},
  pages={15979--15988},
  year={2021}
}

@inproceedings{kakogeorgiou2022hide_attmask,
  title={What to hide from your students: Attention-guided masked image modeling},
  author={Kakogeorgiou, Ioannis and Gidaris, Spyros and Psomas, Bill and Avrithis, Yannis and Bursuc, Andrei and Karantzalos, Konstantinos and Komodakis, Nikos},
  booktitle={Computer Vision--ECCV 2022: 17th European Conference, Tel Aviv, Israel, October 23--27, 2022, Proceedings, Part XXX},
  pages={300--318},
  year={2022},
  organization={Springer}
}

@InProceedings{XieCVPR2023_SecretsMIM,
    author    = {Xie, Zhenda and Geng, Zigang and Hu, Jingcheng and Zhang, Zheng and Hu, Han and Cao, Yue},
    title     = {Revealing the Dark Secrets of Masked Image Modeling},
    booktitle = {Proceedings of the IEEE/CVF Conference on Computer Vision and Pattern Recognition (CVPR)},
    month     = {June},
    year      = {2023},
    pages     = {14475-14485}
}

@inproceedings{zhou2022image,
title={Image {BERT} Pre-training with Online Tokenizer},
author={Jinghao Zhou and Chen Wei and Huiyu Wang and Wei Shen and Cihang Xie and Alan Yuille and Tao Kong},
booktitle={International Conference on Learning Representations (ICLR)},
year={2022}
}

@inproceedings{jiang2022self_SMIT,
  title={Self-supervised 3D anatomy segmentation using self-distilled masked image transformer (SMIT)},
  author={Jiang, Jue and Tyagi, Neelam and Tringale, Kathryn and Crane, Christopher and Veeraraghavan, Harini},
  booktitle={Medical Image Computing and Computer Assisted Intervention--MICCAI 2022: 25th International Conference, Singapore, September 18--22, 2022, Proceedings, Part IV},
  pages={556--566},
  year={2022},
  organization={Springer}
}

@inproceedings{bao2021beit,
title={{BEiT}: {BERT} Pre-Training of Image Transformers},
author={Hangbo Bao and Li Dong and Songhao Piao and Furu Wei},
booktitle={International Conference on Learning Representations (ICLR)},
year={2022},
url={https://openreview.net/forum?id=p-BhZSz59o4}
}

@inproceedings{liu2023patchdropout,
  title={Patchdropout: Economizing vision transformers using patch dropout},
  author={Liu, Yue and Matsoukas, Christos and Strand, Fredrik and Azizpour, Hossein and Smith, Kevin},
  booktitle={Proceedings of the IEEE/CVF Winter Conference on Applications of Computer Vision},
  pages={3953--3962},
  year={2023}
}

@inproceedings{tang2022self,
  title={Self-supervised pre-training of swin transformers for 3d medical image analysis},
  author={Tang, Yucheng and Yang, Dong and Li, Wenqi and Roth, Holger R and Landman, Bennett and Xu, Daguang and Nath, Vishwesh and Hatamizadeh, Ali},
  booktitle={Proceedings of the IEEE/CVF Conference on Computer Vision and Pattern Recognition},
  pages={20730--20740},
  year={2022}
}

@article{tarvainen2017mean,
  title={Mean teachers are better role models: Weight-averaged consistency targets improve semi-supervised deep learning results},
  author={Tarvainen, Antti and Valpola, Harri},
  journal={Advances in neural information processing systems},
  volume={30},
  year={2017}
}

@inproceedings{liu2021swin,
  title={Swin transformer: Hierarchical vision transformer using shifted windows},
  author={Liu, Ze and Lin, Yutong and Cao, Yue and Hu, Han and Wei, Yixuan and Zhang, Zheng and Lin, Stephen and Guo, Baining},
  booktitle={Proceedings of the IEEE/CVF international conference on computer vision},
  pages={10012--10022},
  year={2021}
}

@article{li2021mst,
  title={Mst: Masked self-supervised transformer for visual representation},
  author={Li, Zhaowen and Chen, Zhiyang and Yang, Fan and Li, Wei and Zhu, Yousong and Zhao, Chaoyang and Deng, Rui and Wu, Liwei and Zhao, Rui and Tang, Ming and others},
  journal={Advances in Neural Information Processing Systems},
  volume={34},
  pages={13165--13176},
  year={2021}
}

@inproceedings{caron2021emerging,
  title={Emerging properties in self-supervised vision transformers},
  author={Caron, Mathilde and Touvron, Hugo and Misra, Ishan and J{\'e}gou, Herv{\'e} and Mairal, Julien and Bojanowski, Piotr and Joulin, Armand},
  booktitle={Proceedings of the IEEE/CVF International Conference on Computer Vision},
  pages={9650--9660},
  year={2021}
}

@inproceedings{LiuAAAI2023_AttentionThrow,
  title={Good helper is around you: Attention-driven masked image modeling},
  author={Liu, Zhengqi and Gui, Jie and Luo, Hao},
  booktitle={Proceedings of the AAAI Conference on Artificial Intelligence},
  pages={1799--1807},
  year={2023}
}

@inproceedings{shi2022adversarial_mask,
  title={Adversarial masking for self-supervised learning},
  author={Shi, Yuge and Siddharth, N and Torr, Philip and Kosiorek, Adam R},
  booktitle={International Conference on Machine Learning},
  pages={20026--20040},
  year={2022},
  organization={PMLR}
}

@article{li2022semmae,
  title={Semmae: Semantic-guided masking for learning masked autoencoders},
  author={Li, Gang and Zheng, Heliang and Liu, Daqing and Wang, Chaoyue and Su, Bing and Zheng, Changwen},
  journal={Advances in Neural Information Processing Systems},
  volume={35},
  pages={14290--14302},
  year={2022}
}

@article{lee2023self,
  title={Self-knowledge distillation via dropout},
  author={Lee, Hyoje and Park, Yeachan and Seo, Hyun and Kang, Myungjoo},
  journal={Computer Vision and Image Understanding},
  volume={233},
  pages={103720},
  year={2023},
  publisher={Elsevier}
}

@inproceedings{laine2016temporal,
  title={Temporal ensembling for semi-supervised learning},
  author={Laine, Samuli and Aila, Timo},
  booktitle={International Conference on Learning Representations (ICLR)},
  year={2017}
}

@book{xiao2023lesion,
  title={Lesion Segmentation in Surgical and Diagnostic Applications: MICCAI 2022 Challenges, CuRIOUS 2022, KiPA 2022 and MELA 2022, Held in Conjunction with MICCAI 2022, Singapore, September 18--22, 2022, Proceedings},
  author={Xiao, Yiming and Yang, Guanyu and Song, Shuang},
  volume={13648},
  year={2023},
  publisher={Springer Nature}
}

@article{ji2022amos,
  title={Amos: A large-scale abdominal multi-organ benchmark for versatile medical image segmentation},
  author={Ji, Yuanfeng and Bai, Haotian and Yang, Jie and Ge, Chongjian and Zhu, Ye and Zhang, Ruimao and Li, Zhen and Zhang, Lingyan and Ma, Wanling and Wan, Xiang and others},
  journal={arXiv preprint arXiv:2206.08023},
  year={2022}
}

@misc{aerts2015data,
  title={Data from \textsc{NSCLC}-radiomics. \textsc{T}he \textsc{C}ancer \textsc{I}maging \textsc{A}rchive},
  author={Aerts, H. and Rios V. E. and Leijenaar, Ralph TH and Parmar, C. and Grossmann, P. and Carvalho, S. and Lambin, P.},
  year={2015}
}

@inproceedings{cao2022swin,
  title={Swin-unet: Unet-like pure transformer for medical image segmentation},
  author={Cao, Hu and Wang, Yueyue and Chen, Joy and Jiang, Dongsheng and Zhang, Xiaopeng and Tian, Qi and Wang, Manning},
  booktitle={European conference on computer vision},
  pages={205--218},
  year={2022},
  organization={Springer}
}

@article{mcinnes2018umap, doi = {10.21105/joss.00861}, url = {https://doi.org/10.21105/joss.00861}, year = {2018}, publisher = {The Open Journal}, volume = {3}, number = {29}, pages = {861}, author = {Leland McInnes and John Healy and Nathaniel Saul and Lukas Großberger}, title = {UMAP: Uniform Manifold Approximation and Projection}, journal = {Journal of Open Source Software} }

@article{zhu2023wasserstein,
  title={Wasserstein HOG: Local Directionality Extraction via Optimal Transport},
  author={Zhu, Jiening and Veeraraghavan, Harini and Jiang, Jue and Oh, Jung Hun and Norton, Larry and Deasy, Joseph O and Tannenbaum, Allen},
  journal={IEEE transactions on medical imaging},
  year={2023},
  publisher={IEEE}
}

@inproceedings{Paszke2019PyTorchAI,
  title={PyTorch: An Imperative Style, High-Performance Deep Learning Library},
  author={Adam Paszke and Sam Gross and Francisco Massa and Adam Lerer and James Bradbury and Gregory Chanan and Trevor Killeen and Zeming Lin and Natalia Gimelshein and Luca Antiga and Alban Desmaison and Andreas K{\"o}pf and Edward Yang and Zach DeVito and Martin Raison and Alykhan Tejani and Sasank Chilamkurthy and Benoit Steiner and Lu Fang and Junjie Bai and Soumith Chintala},
  booktitle={Neural Information Processing Systems},
  year={2019},
  url={https://api.semanticscholar.org/CorpusID:202786778}
}

@article{isensee2021nnu,
  title={nnU-Net: a self-configuring method for deep learning-based biomedical image segmentation},
  author={Isensee, Fabian and Jaeger, Paul F and Kohl, Simon AA and Petersen, Jens and Maier-Hein, Klaus H},
  journal={Nature methods},
  volume={18},
  number={2},
  pages={203--211},
  year={2021},
  publisher={Nature Publishing Group}
}

@inproceedings{arani_FickleTeacherWACV2021,
  title={Noise as a resource for learning in knowledge distillation},
  author={Arani, Elahe and Sarfraz, Fahad and Zonooz, Bahram},
  booktitle={Proceedings of the IEEE/CVF Winter Conference on Applications of Computer Vision},
  pages={3129--3138},
  year={2021}
}

@inproceedings{liu2020noisy,
    title = "Noisy Self-Knowledge Distillation for Text Summarization",
    author = "Liu, Yang  and
      Shen, Sheng  and
      Lapata, Mirella",
    editor = "Toutanova, Kristina  and
      Rumshisky, Anna  and
      Zettlemoyer, Luke  and
      Hakkani-Tur, Dilek  and
      Beltagy, Iz  and
      Bethard, Steven  and
      Cotterell, Ryan  and
      Chakraborty, Tanmoy  and
      Zhou, Yichao",
    booktitle = "Proceedings of the 2021 Conference of the North American Chapter of the Association for Computational Linguistics: Human Language Technologies",
    month = jun,
    year = "2021",
    publisher = "Association for Computational Linguistics"
}

@article{yang2023medmnist,
  title={MedMNIST v2-A large-scale lightweight benchmark for 2D and 3D biomedical image classification},
  author={Yang, Jiancheng and Shi, Rui and Wei, Donglai and Liu, Zequan and Zhao, Lin and Ke, Bilian and Pfister, Hanspeter and Ni, Bingbing},
  journal={Scientific Data},
  volume={10},
  number={1},
  pages={41},
  year={2023},
  publisher={Nature Publishing Group UK London}
}

@inproceedings{bulo2016dropout,
  title={Dropout distillation},
  author={Bul{\`o}, Samuel Rota and Porzi, Lorenzo and Kontschieder, Peter},
  booktitle={International Conference on Machine Learning},
  pages={99--107},
  year={2016},
  organization={PMLR}
}

@article{you2022simcvd,
  title={Simcvd: Simple contrastive voxel-wise representation distillation for semi-supervised medical image segmentation},
  author={You, Chenyu and Zhou, Yuan and Zhao, Ruihan and Staib, Lawrence and Duncan, James S},
  journal={IEEE Transactions on Medical Imaging},
  volume={41},
  number={9},
  pages={2228--2237},
  year={2022},
  publisher={IEEE}
}

@inproceedings{wang2022bevt,
  title={Bevt: Bert pretraining of video transformers},
  author={Wang, Rui and Chen, Dongdong and Wu, Zuxuan and Chen, Yinpeng and Dai, Xiyang and Liu, Mengchen and Jiang, Yu-Gang and Zhou, Luowei and Yuan, Lu},
  booktitle={Proceedings of the IEEE/CVF conference on computer vision and pattern recognition},
  pages={14733--14743},
  year={2022}
}

@inproceedings{dosovitskiy2021,
title={An Image is Worth 16x16 Words: Transformers for Image Recognition at Scale},
author={Alexey Dosovitskiy and Lucas Beyer and Alexander Kolesnikov and Dirk Weissenborn and Xiaohua Zhai and Thomas Unterthiner and Mostafa Dehghani and Matthias Minderer and Georg Heigold and Sylvain Gelly and Jakob Uszkoreit and Neil Houlsby},
booktitle={International Conference on Learning Representations (ICLR)},
year={2021}
}

@inproceedings{wei2022masked_feature,
  title={Masked feature prediction for self-supervised visual pre-training},
  author={Wei, Chen and Fan, Haoqi and Xie, Saining and Wu, Chao-Yuan and Yuille, Alan and Feichtenhofer, Christoph},
  booktitle={Proceedings of the IEEE/CVF Conference on Computer Vision and Pattern Recognition},
  pages={14668--14678},
  year={2022}
}

@inproceedings{xie2021simmim,
  title={Simmim: A simple framework for masked image modeling},
  author={Xie, Zhenda and Zhang, Zheng and Cao, Yue and Lin, Yutong and Bao, Jianmin and Yao, Zhuliang and Dai, Qi and Hu, Han},
  booktitle={Proceedings of the IEEE/CVF Conference on Computer Vision and Pattern Recognition},
  pages={9653--9663},
  year={2022}
}

@article{cardoso2022monai,
  title={Monai: An open-source framework for deep learning in healthcare},
  author={Cardoso, M Jorge and Li, Wenqi and Brown, Richard and Ma, Nic and Kerfoot, Eric and Wang, Yiheng and Murrey, Benjamin and Myronenko, Andriy and Zhao, Can and Yang, Dong and others},
  journal={arXiv preprint arXiv:2211.02701},
  year={2022}
}

@inproceedings{Loshchilov2017DecoupledWD,
  title={Decoupled Weight Decay Regularization},
  author={Ilya Loshchilov and Frank Hutter},
  booktitle={International Conference on Learning Representations},
  year={2017},
  url={https://api.semanticscholar.org/CorpusID:53592270}
}

@article{armato2011lung,
  title={The lung image database consortium (LIDC) and image database resource initiative (IDRI): a completed reference database of lung nodules on CT scans},
  author={Armato III, Samuel G and McLennan, Geoffrey and Bidaut, Luc and McNitt-Gray, Michael F and Meyer, Charles R and Reeves, Anthony P and Zhao, Binsheng and Aberle, Denise R and Henschke, Claudia I and Hoffman, Eric A and others},
  journal={Medical physics},
  volume={38},
  number={2},
  pages={915--931},
  year={2011},
  publisher={Wiley Online Library}
}

@article{chen2019med3d,
  title={Med3d: Transfer learning for 3d medical image analysis},
  author={Chen, Sihong and Ma, Kai and Zheng, Yefeng},
  journal={arXiv preprint arXiv:1904.00625},
  year={2019}
}

@article{Loshchilov2016SGDRSG,
  title={SGDR: Stochastic Gradient Descent with Warm Restarts},
  author={Ilya Loshchilov and Frank Hutter},
  journal={arXiv: Learning},
  year={2016},
  url={https://api.semanticscholar.org/CorpusID:14337532}
}
	
	\clearpage
	\appendix

	\section{Implementation details}
	\label{sup_subsec:implementationdetails}
	
	Pytorch~\cite{Paszke2019PyTorchAI} and MONAI~\cite{cardoso2022monai} libraries were used for implementation and training of the various models. For pretraining, DAGMaN and various comparable methods such as AttMask~\cite{kakogeorgiou2022hide_attmask}, MST~\cite{li2021mst}, iBot~\cite{zhou2022image} and SMIT~\cite{jiang2022self_SMIT} were pretrained on datasets as mentioned in Section \ref{subsec:pretrainingdatasets}.
	
	Two different transformer structures were evaluated in this study, namely ViT and Swin. The ViT~\cite{dosovitskiy2021} architecture comprised of 12 transformer blocks, 768 embedding features, and 8 multi-head self attention. The Swin~\cite{liu2021swin} architecture used a depth of [2,2,8,2] and [4,4,8,16] multi-head for each transformer depth, and a feature embedding size of 384. This setup also included a window size of 4 $\times$ 4 $\times$ 4 and patch size of 2 $\times$ 2 $\times$ 2.
	
	We generated augmented views by re-sampling the scans at 2mm $\times$ 2mm $\times$ 2mm voxel spacing and then randomly cropping 128 $\times$ 128 $\times$ 128 voxel scans. The networks were optimized using ADAMw~\cite{Loshchilov2017DecoupledWD} with a cosine learning rate scheduler~\cite{Loshchilov2016SGDRSG} trained for 800 epochs with an initial learning rate of $8e^{-4}$ and warmup for 80 epochs. A path drop rate of 0.1 was applied to the student model, and pretraining was conducted on four NVIDIA A100 GPUs (each with 80GB memory). Hyperparameters $\lambda_{AITD}$ = 0.1, $\lambda_{GITD}$ = 0.1, amd $\lambda_{AMPD}$ = 0.1 in Section \ref{subsec:noisyteacher} were determined experimentally via grid-search. Degenerate solutions were avoided using centering and sharpening operations~\cite{caron2021emerging,jiang2022self_SMIT}. 
	
	\begin{figure}[b]
		\centering
		\includegraphics[width=1.0\textwidth]{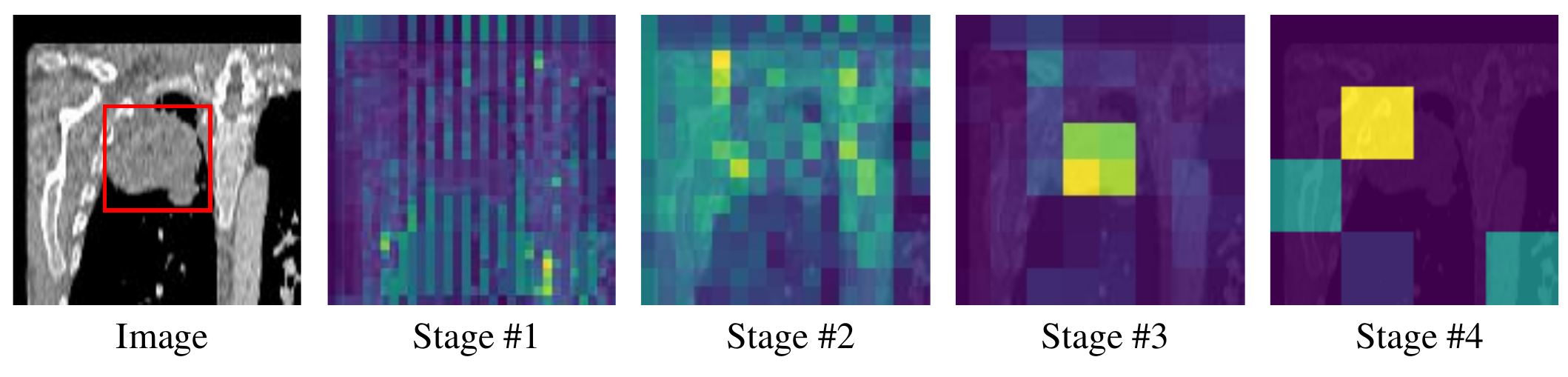} 
		\caption{The pretrained attention map when putting the semantic attention layers at different stage of Swin transformer. }
		\label{fig:ablation_attnmap_differentstages_samodule_placement}
	\end{figure}
	
	\section{Downstream task descriptions}
	\label{sup_subsec:downsteamdatadescriptions}
	
	\noindent\textbf{Task 1: Binary classification of Malignant and Benign lung nodules from patients screened for lung cancer (LC)}
	
	The Lung Image Database Consortium (LIDC) dataset~\cite{armato2011lung} consists of 1,010 patients from seven different institutions, with a total of 2,426 lung nodules extracted using pylidc library. Each nodule was rated on a malignancy from scale 1-5, with ratings of 1$-$3 as benign and 4$-$5 as malignant, following the approach of~\cite{chen2019med3d}. This grouping approach resulted in 2,054 benign and 540 malignant nodules. We employed a 3-fold stratified cross-validation on 1624 nodules, and used the rest 1,000 as test set. We used a batch size of 40 with a learning rate of $2e^{-4}$ for 1,000 epochs on NVIDIA A40 for fine-tuning and evaluated the models using the Area Under the Curve (AUC) metric.
	\\
	
	\noindent\textbf{Task 2: Lung tumor and nodule segmentation} 
	
	For segmentation, we used 377 3D CT scans from lung cancer (LC) patients before radiation treatment, sourced from the publicly available cancer imaging archive (TCIA-LC)~\cite{aerts2015data} dataset. We fine-tuned our model on this dataset (350 for training and 27 used for validation), and then tested on the LIDC containing lung nodules~\cite{armato2011lung}. As this data represents nodules that are typically small compared to malignant cancers, results for nodules larger than 3 cc were reported. 
	In addition, the model was also evaluated in an institutional dataset of 200 patients with advanced non-small cell lung cancers (same patients used in Task 3 below).
	\\
	
	\begin{figure}[t]
		\def\arraystretch{1.2}
		\scriptsize
		\centering
		\begin{minipage}{0.45\textwidth} 
			\centering
			\includegraphics[width=1.0\textwidth]{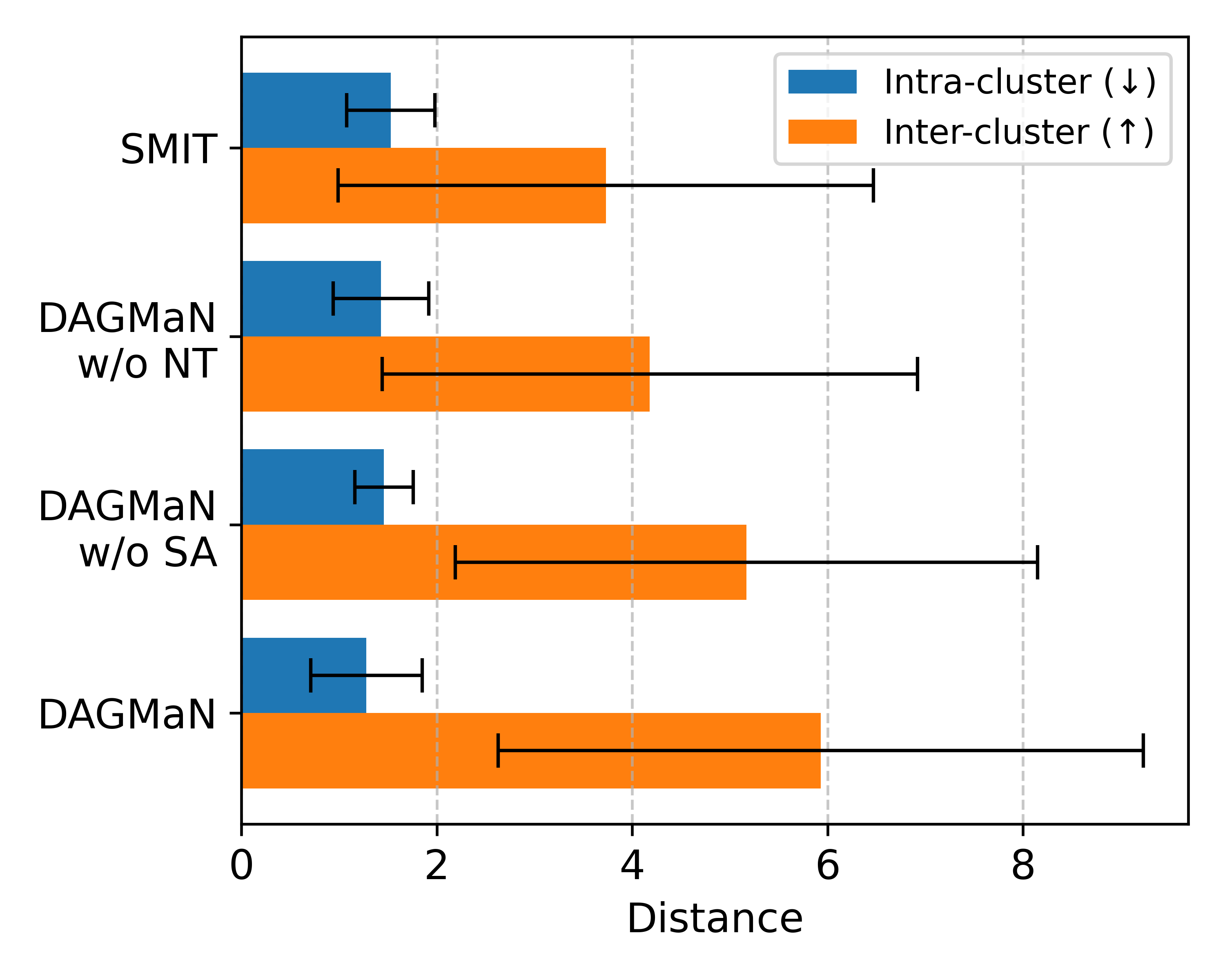} 
			\caption{The impact noisy teacher and semantic attention on the inter- and intra cluster of the OrganMNIST dataset.}
			\label{fig:results_cluster_distance}
		\end{minipage}
		\hfill
		\begin{minipage}{0.5\textwidth} 
			\centering
			\captionof{table}{Impact of losses on nodule malignancy classification}
			\label{tab:ablation_differentlosses}
			\resizebox{\textwidth}{!}{%
				\begin{tabular}{lllll}
					AMIP & AMPD & AITD & GITD & AUC \\ \shline
					$\checkmark$ & $\times$ & $\times$ & $\times$ & 0.859\\
					$\checkmark$ & $\checkmark$ & $\times$ & $\times$ & 0.869 \\
					$\checkmark$ & $\checkmark$ & $\checkmark$ & $\times$ & 0.878 \\
					$\checkmark$ & $\times$ & $\checkmark$ & $\checkmark$ & 0.865 \\
					$\times$ & $\checkmark$ & $\checkmark$ & $\checkmark$ & 0.829 \\
					$\checkmark$ & $\checkmark$ & $\checkmark$ & $\checkmark$ & 0.895 \\ \shline
				\end{tabular}%
			}
		\end{minipage}
		
		\hfill
		\begin{minipage}{0.48\textwidth} 
			\centering
			\includegraphics[width=1.0\textwidth]{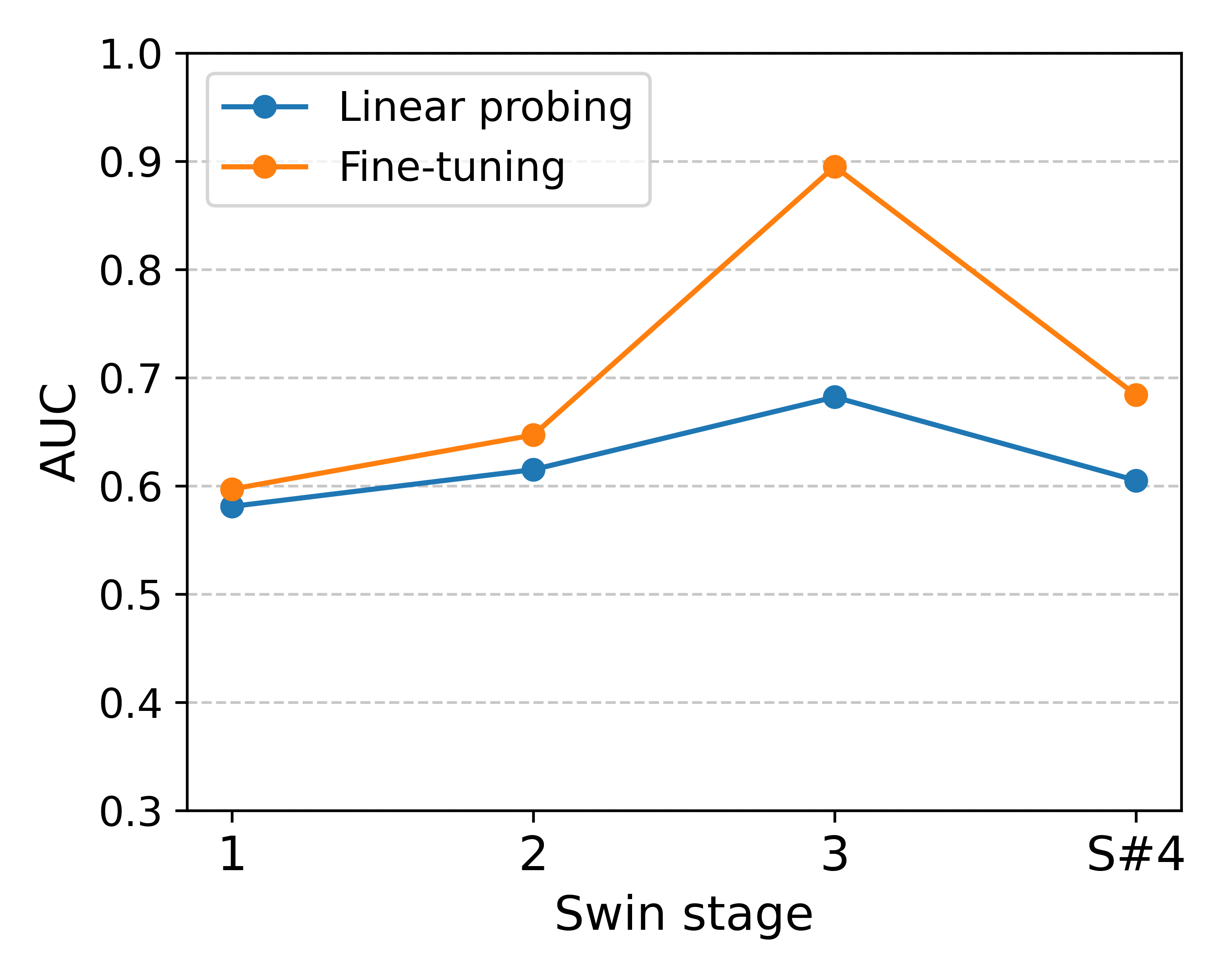} 
            \caption{Impact of semantic attention module placement on nodule malignancy classification}
			\label{fig:ablation_lidc_samodule_placement}
		\end{minipage}
		\hfill
		\begin{minipage}{0.48\textwidth}
			\centering
			\includegraphics[width=1.0\textwidth]{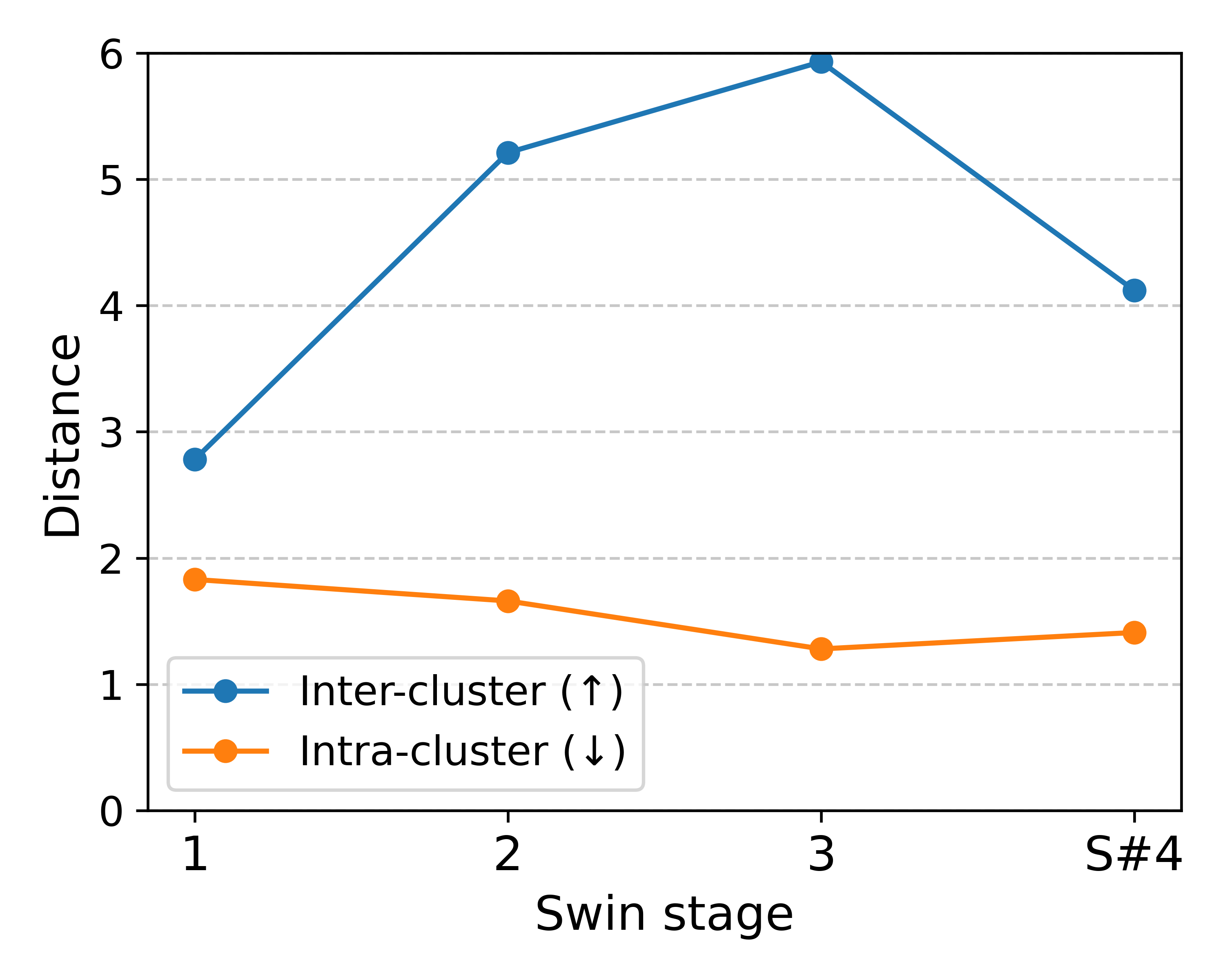} 
			\caption{Impact of semantic attention module placement on unsupervised organ clustering}
			\label{fig:ablation_clustering_samodule_placement}
		\end{minipage}
	\end{figure}
	
	\noindent\textbf{Task 3: 
		Binary prediction of immunotherapy response}
	
	We defined response no disease progression $\geq$ 6 months after start of immunotherapy treatment start vs disease progression within 6 months as non durable benefit [NDB]). Pre-treatment CT scans of 200 patients with non-small cell lung cancers were used. Patients were scanned either with contrast or non-contrast scans and acquired using lung reconstruction kernel. The response distribution of Durable Clinical Benefit [DCB] to Non-Durable Benefit [NDB] is 82:118. Three-fold stratified cross-validation was applied.  We used a batch size of 40 with a learning rate of $2e^{-4}$ for 1,000 epochs on NVIDIA A40 for fine-tuning and evaluated the models using the Area Under the Curve (AUC) metric. 
	\\
	\\
	\noindent\textbf{Task 4: 
		Unsupervised clustering of organs from OrganMNIST}
	
	We evaluated our pretrained model's capability to differentiate various organs on the OrganMNIST3D dataset, which consists of 1,743 3D CT scans representing 11 different organs in the abdomen. The performance was evaluated using inter-cluster and intra-cluster distances, computed using the Euclidean distance.
	
	\begin{figure}
		\centering
		\includegraphics[width=0.95\linewidth]{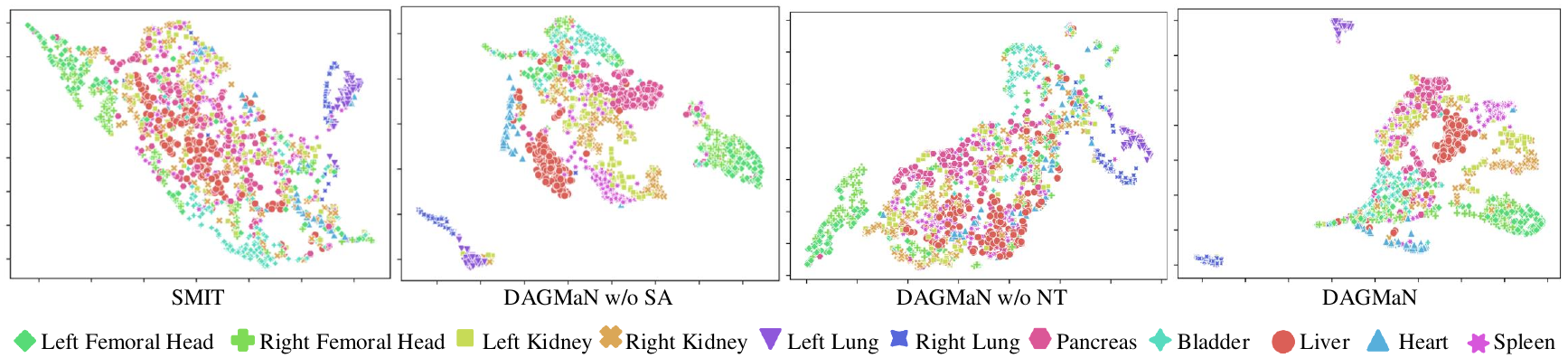}
		\caption{Differences in unsupervised clustering of distinct organs using pretrained features from different configurations.}
		\label{fig:umap_hq}
	\end{figure}

	\begin{table}[]
		\def\arraystretch{1.2}
		\scriptsize
		\centering
		\caption{DAGMaN generalized to both ViT and Swin transformer.}
		\label{tab:ablation_performance_vitswin}
		\resizebox{\columnwidth}{!}{%
			\begin{tabular}{ll|ll|ll}
				\multicolumn{2}{c|}{} & \multicolumn{2}{c|}{Nodule malignancy classification} & \multicolumn{2}{c}{Unsupervised organ clustering} \\ 
				SSL Task & Network & Linear probing & Fine-tuning & Intra-cluster ($\downarrow$) & Inter-cluster ($\uparrow$) \\ \shline
				SMIT & ViT & 0.643 & 0.816 & 1.30 $\pm$ 0.51 & 6.44 $\pm$ 5.27 \\
				\hspace{0.5em}DAGMaN & ViT & 0.662 & 0.847 & 1.15 $\pm$ 0.41 & 6.59 $\pm$ 4.60 \\
				SMIT & Swin & 0.629 & 0.859 & 1.53 $\pm$ 0.45 & 3.73 $\pm$ 2.74 \\
				\hspace{0.5em}DAGMaN & Swin & 0.642 & 0.895 & 1.28 $\pm$ 0.57 & 5.93 $\pm$ 3.30 \\ 
				\hline
			\end{tabular}%
		}
	\end{table}

\end{document}